\documentclass{article} 
\usepackage{iclr2021_conference,times}

\usepackage{amsmath,amsfonts,bm}

\def\eqref#1{equation~\ref{#1}}

\def\1{\bm{1}}

\DeclareMathAlphabet{\mathsfit}{\encodingdefault}{\sfdefault}{m}{sl}
\SetMathAlphabet{\mathsfit}{bold}{\encodingdefault}{\sfdefault}{bx}{n}

\usepackage{hyperref}
\usepackage{url}

\usepackage{svg}
\usepackage{graphicx} 
\usepackage{float} 
\usepackage{subfigure} 
\usepackage{tcolorbox}
\usepackage{CJKutf8}
\definecolor{dt}{gray}{0.7}
\usepackage{multicol}
\usepackage{multirow}
\usepackage{booktabs}
\usepackage{multirow}
\usepackage{makecell}
\usepackage{xspace}
\DeclareSymbolFont{extraup}{U}{zavm}{m}{n}
\DeclareMathSymbol{\varheart}{\mathalpha}{extraup}{86}
\DeclareMathSymbol{\vardiamond}{\mathalpha}{extraup}{87}

\newcommand{\expec}{\mathbb{E}}
\newcommand{\encoder}{\mathcal{E}}

\newcommand{\zt}[1]{z_{#1}}

\newcommand{\model}{\epsilon_\theta}
\newcommand{\conditioner}{\tau_\theta}
\usepackage{tablefootnote}

\title{\Large Taiyi-Diffusion-XL: Advancing Bilingual Text-to-Image Generation with Large Vision-Language Model Support}

\author{
    Xiaojun Wu$^{\varheart}$\footnotemark[1] \qquad
    Dixiang Zhang$^{\varheart\vardiamond}$\footnotemark[1] \qquad
    Ruyi Gan$^{\varheart\clubsuit}$\footnotemark[2] \qquad
    \\
\hspace{8mm}\textbf{
    Junyu Lu$^{\varheart}$\ \qquad
    Ziwei Wu$^{\varheart}$\ \qquad
    Renliang Sun$^{\varheart}$\ \qquad
    }
    \\
\hspace{8mm}\textbf{
    Jiaxing Zhang$^{\varheart}$\footnotemark[3] \qquad
    Pingjian Zhang$^{\vardiamond}$\footnotemark[3] \qquad
    Yan Song$^{\clubsuit}$\footnotemark[3] \qquad
    }
    \\
    \\
\hspace{2mm}$^\varheart$International Digital Economy Academy \quad
    $^\clubsuit$South China University of Technology \quad \\
    \hspace{-2mm}$^\vardiamond$University of Science and Technology of China \quad 
    \\
    \\
    \hspace{2mm}{\tt\small \{wuxiaojun, zhangdixiang, ganruyi, lujunyu, wuziwei, sunrenliang\}@idea.edu.cn } \\
    {\tt\small zhangjiaxing@idea.edu.cn} \hspace{4mm}
    {\tt\small pjzhang@scut.edu.cn}\hspace{4mm}
    {\tt\small clksong@gmail.com}
}

\begin{document}

\maketitle

{
  \renewcommand{\thefootnote}%
    {\fnsymbol{footnote}}
  \footnotetext[1]{Equal Contribution.}
  \footnotetext[2]{Project Leader.}
  \footnotetext[3]{Corresponding Author.}
}

\begin{abstract}
Recent advancements in text-to-image models have significantly enhanced image generation capabilities, yet a notable gap of open-source models persists in bilingual or Chinese language support.
To address this need, we present Taiyi-Diffusion-XL, a new Chinese and English bilingual text-to-image model which is developed by extending the capabilities of CLIP and Stable-Diffusion-XL through a process of bilingual continuous pre-training.
This approach includes the efficient expansion of vocabulary by integrating the most frequently used Chinese characters into CLIP's tokenizer and embedding layers, coupled with an absolute position encoding expansion.
Additionally, we enrich text prompts by large vision-language model, leading to better images captions and possess higher visual quality.
These enhancements are subsequently applied to downstream text-to-image models. Our empirical results indicate that the developed CLIP model excels in bilingual image-text retrieval.
Furthermore, the bilingual image generation capabilities of Taiyi-Diffusion-XL surpass previous models.
This research leads to the development and open-sourcing of the Taiyi-Diffusion-XL model, representing a notable advancement in the field of image generation, particularly for Chinese language applications.
%
%
The model and demonstration are made publicly available at ~\url{https://huggingface.co/IDEA-CCNL/Taiyi-Stable-Diffusion-XL-3.5B/}, fostering further research and collaboration in this domain.
\end{abstract}
\section{Introduction}

Recent developments in diffusion models, such as those presented in Stable Diffusion(SD) \citep{Rombach_2022_CVPR_sd,podell2023sdxl},DALL-E \citep{ramesh2022dalle2, james2022dalle3}, Imagen \citep{saharia2022imagen}, and Deepfloyd-IF \citep{alex2023deep-floyd_IF}, have showcased their potential in generating high-quality images from text descriptions.
However, it is important to note that the majority of current open-source text-to-image models predominantly support English, with very few offering bilingual support for both Chinese and English.
This advancement diverges from the conventional methodology of employing translation software to convert Chinese text into English for subsequent image generation with English-centric models.
In particular, works such as Taiyi-Diffusion \citep{zhang2022fengshenbang}, Pai-Diffusion \citep{wang2023pai} and Alt-Diffusion\citep{Ye2023AltDiffusionAM} have made significant strides in adapting text-to-image models for Chinese scenarios, demonstrating the feasibility and importance of native language support in such models. 
Such models adeptly handle language-specific expressions, ensuring the preservation of original meaning and emotional nuances that might otherwise be lost in translation process. 
These models often obtain Chinese understanding capabilities by replacing multi-language text encoders \citep{radford2021clip,devlin2019bert} and retaining unet\citep{ronneberger2015unet} while this methodology will discard the original English understanding capabilities.

\begin{figure*}[htbp]
    \centering
    \includegraphics[width=\linewidth]{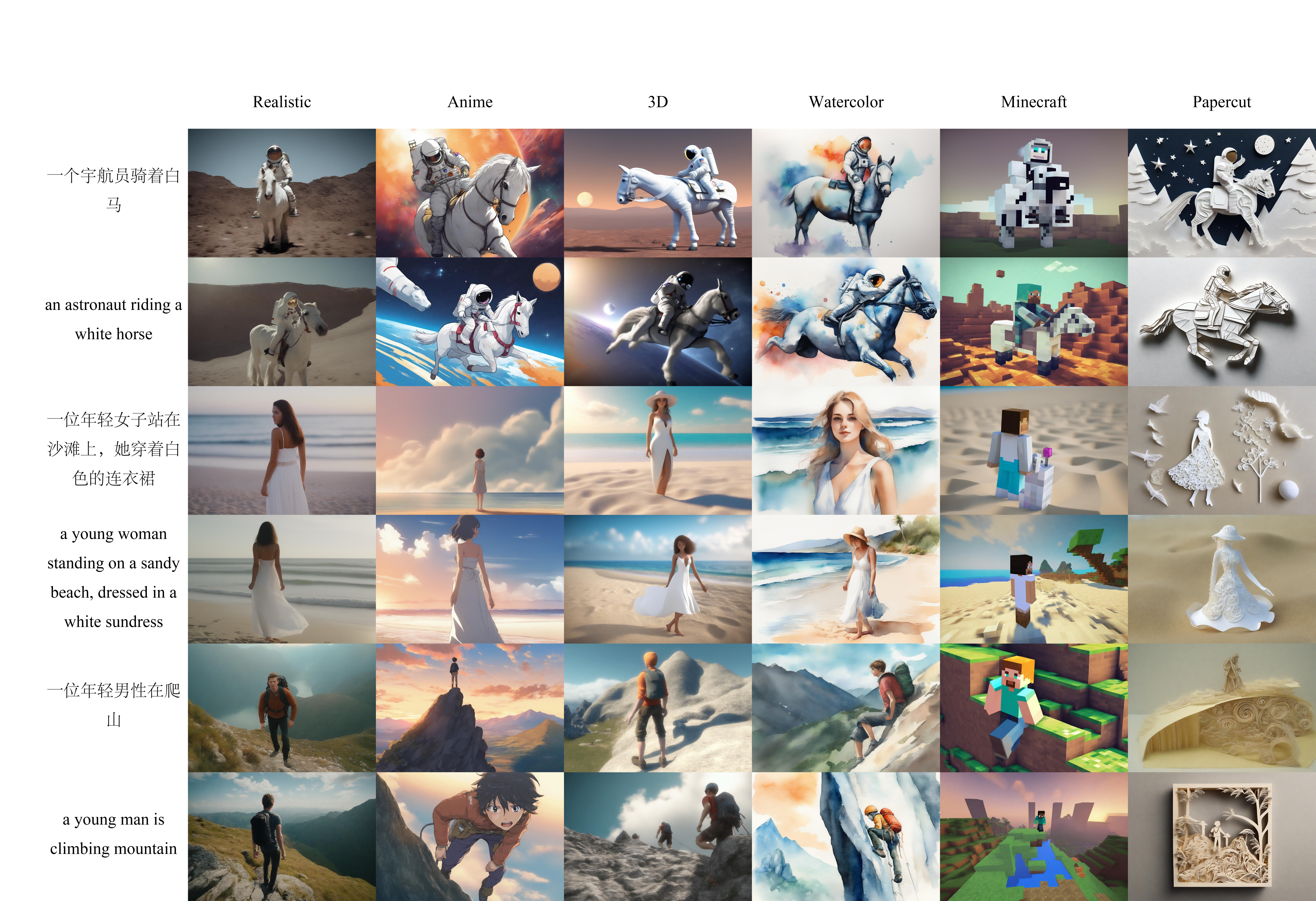}
    \caption{
    An illustration of Taiyi-XL showcasing text-to-image generation results under various styles and prompts.
    }
    \label{fig: high-resolution}
\end{figure*}

Building on these advancements, our work, Taiyi-Diffusion-XL(Taiyi-XL), specifically focuses on augmenting these models for Chinese text-to-image generation while preserving the original English ability, addressing the unique linguistic and cultural aspects of the bilingual language.
In summary, while translation tools offer a certain level of convenience for cross-language applications, native language support in models, especially for languages like Chinese, provides distinct advantages in terms of comprehension, accuracy, and efficiency. Our contributions are aimed at enhancing these capabilities, thereby offering more effective and inclusive tools for the research community. Our research contributes to this evolving field in three significant ways:

\begin{itemize}

    \item \textit{Efficient Algorithms for Bilingual Expansion}: We develop algorithms for expanding vocabulary and position encoding in text-to-image models tailored for bilingual contexts. This advancement facilitates more accurate and culturally tuned image generation.

    \item \textit{Enrichment of Text Prompts by Large Vision-Language Models}: We employ large vision-language models to enrich text prompts. This approach marks a substantial enhancement in the model's ability to interpret and visualize complex textual descriptions.
    
    \item \textit{Creation of Bilingual Models}: Utilizing the capabilities of multimodal foundation model, we develop and open-source the text-to-image model, Taiyi-XL, which significantly advances the research and application of bilingual text-to-image models.

\end{itemize}

\section{Methodology}

Our methodology for text-to-image generation, especially with diffusion models, encompasses two primary phases, focusing on dataset preparation and model training.

\subsection{Dataset Preparation}

We curate a dataset consisting of high-quality image-text pairs $(X, Y)$, where $X$ represents an image, and $Y$ is a descriptive text. In contrast to traditional datasets with discretized tags, our dataset emphasizes comprehensive descriptions, capturing materials, styles, colors, and spatial layouts. To address the limitations of web-crawled resources, which often contain irrelevant or inaccurate tags, we employ vision-language large models \citep{lu2023ziyavl,lu2023lyrics} to generate synthetic captions that more accurately describe the images, which inherits the language capabilities of the bilingual large language model \citep{gan2023ziya2} and expands the visual capabilities of LLMs. This approach not only enhances the richness of the dataset but also ensures a higher degree of relevance and detail in the descriptions. We use images, web crawl caption, and instructions for generating description as inputs for the Lyrics \citep{lu2023lyrics}. In Chinese, we select \begin{CJK*}{UTF8}{gbsn}``请详细描述图片内容。"\end{CJK*} as the instruction, and in English, we select ``Write a detailed description of the given image." as the instruction. The Lyrics model generates new, accurate descriptive text by extracting features from the images as well as distilling useful information from inaccurate and imperfect web crawl captions. Finally, we combine the generated high-quality text with the original images to form image-text pairs, which are then input into the Taiyi-XL for training.

\subsection{CLIP Training}
\label{subsec: clip_train}
The foundation of our model is a vision-language large model, similar to CLIP~\citep{radford2021clip}, which aligns images and text representations effectively. We start with the pre-trained English-only CLIP model and extend its training to accommodate bilingual adaptation and the nuanced requirements of high-quality image-text data. The first stage of training involves processing a large-scale, bilingual dataset, including Laion \citep{schuhmann2021laion} and Wukong \citep{gu2022wukong}, with a focus on data cleaning and quality enhancement. We employ a contrastive loss function and a distributed, memory-efficient training approach \citep{chen2023discoclip}. The second stage continues with training on our enriched dataset, emphasizing the diverse perspectives and details captured in high-quality image-text pairs.

\subsection{Taiyi-XL Training}
The Taiyi-XL training process, a key component in our text-to-image generation methodology, especially with diffusion models, involves two primary phases:

\begin{figure*}[t]
    \centering
    \includegraphics[width=0.9\linewidth]{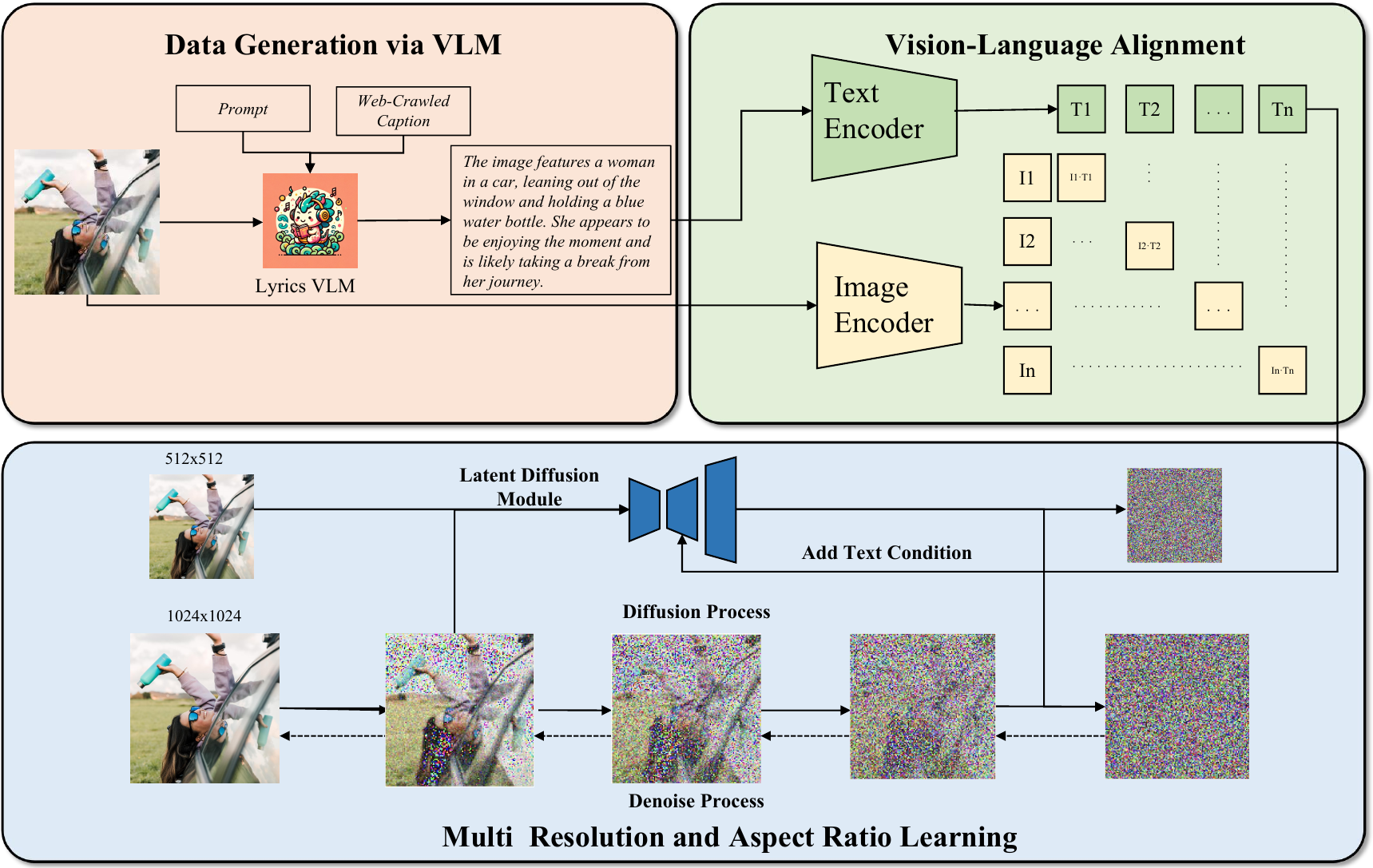}
    \caption{
    Overview of the Taiyi-Diffusion-XL(Taiyi-XL) training process, encompassing data preprocessing, image-text contrastive learning and multi-resolution denoising training process.
    }
    \label{fig:structure}
\end{figure*}

\paragraph{Initialization and Training.}

We initialize the Taiyi-XL model, denoted as $\mathcal{G}_\theta$, with components including a noise predictor $\model$, a CLIP text encoder $\conditioner$ from \ref{subsec: clip_train}, a latent encoder $\encoder$, and a dataset $\mathcal{D}$. Each data instance in $\mathcal{D}$ is represented as a pair \( (x_i, y_i) \), where \( x_i \) is an image and \( y_i \) is its corresponding textual descriptor. For the training phase at mix resolution of \( 512 \times 512 \) and \( 1024 \times 1024 \), we define a loss function \( L \) to guide the image denoising process:
\begin{equation}
 L(\theta) := \expec_{\encoder(x), y, \epsilon \sim \mathcal{N}(0, 1), t }\Big[ \Vert \epsilon - \model(z_{t},t, \conditioner(y)) \Vert_{2}^{2}\Big] \, ,
\label{eq:cond_loss}
\end{equation}
The model is conceptualized as a sequence of denoising latent autoencoders $\model(\zt{t},t);\, t=1\dots T$, implemented as a time-conditional UNet \citep{ronneberger2015unet}. The latent representations $\zt{t}$ are efficiently obtained from $\encoder(x)$ during training and decoded to the image space using a VAE decoder \citep{kingma2013vae}. The text encoder $\conditioner$, parameterized as a transformer model, is optimized jointly with $\model$ as per Eq.~\ref{eq:cond_loss}.

Model parameters \( \theta \) are iteratively updated using gradient descent to minimize the loss function \( L(\theta, e) \):
\begin{equation}
\theta_{e+1} = \theta_e - \eta \cdot \nabla_{\theta}L(\theta_e, e)
\label{eq:backpropagate_loss}
\end{equation}
where \( \eta \) represents the learning rate.

\paragraph{Text-to-Image Generation.}

For text-to-image generation, we utilize the trained bilingual text encoder for extracting features from textual descriptions. The extracted textual features $\conditioner(y)$ are then integrated into the latent diffusion process, enhancing computational efficiency and reducing processing time and memory requirements. In the generation phase, starting from the last time step $T$ with pure noise, the model iteratively denoises the input, converging to \( x_0 \), the clean image, as described by:
\begin{equation} 
x_{t-1} = x_{t} - \epsilon_{\theta}(x_t,t,\conditioner(y)), \quad \lim_{t \to 0} x_t = x_0
\label{eq:inference}
\end{equation}

\section{Experiment And Analysis}
\noindent\textbf{Training Settings.}
We base our Taiyi-XL model on the pre-trained Stable Diffusion XL (SD-XL) \citep{podell2023sdxl} checkpoint, providing a strong foundation for image generation. To enhance efficiency and manage GPU memory use, we adopt the BFLOAT16 format. Our training approach involves a learning rate of 1e-5, starting with a warmup phase for stable learning, followed by a cosine decay schedule to fine-tune and refine the model. These strategies are essential for balancing training speed with model performance.

\noindent\textbf{Evaluation Protocols.}
Our evaluation framework encompasses both machine and human evaluation to provide a comprehensive understanding of the model's performance. Machine evaluation metrics include CLIP performance evaluation with image-to-text retrieval and text-to-image retrieval; CLIP Similarity (CLIP Sim), which measures the semantic alignment between the generated images and text descriptions; Inception Score (IS), assessing the quality and diversity of the images; and Fréchet Inception Distance (FID), evaluating the distance between the distributions of generated and real images. 
In the context of human evaluation of text-to-image generation, it is acknowledged that such assessments inherently possess a degree of subjectivity. Consequently, this study primarily employs a case analysis approach to discern and articulate the distinct characteristics of image generation outcomes produced by different models. Rather than providing direct quantitative results that delineate superiority or inferiority among the models, the focus is on a qualitative examination that highlights the unique attributes and performance nuances of each model in image generation tasks. 

\noindent\textbf{Baselines.} 
For our comparative analysis, we include several established models as baselines: SD-XL \citep{podell2023sdxl}, Midjourney\footnote{https://www.midjourney.com/}, DALL-E 3\footnote{https://openai.com/dall-e-3} \citep{james2022dalle3}, along with other open-sourced models such as our previsous work Taiyi-v0.1\citep{fengshenbang}, Alt-Diffusion\citep{Ye2023AltDiffusionAM} and Pai-Diffusion\citep{wang2023pai}. DALL-E 3, recognized for its innovative text-to-image capabilities, sets a high standard in generating quality images from text descriptions. SD-XL, a variant of the Stable Diffusion model, excels in complex image synthesis tasks. By comparing Taiyi-XL with these models, we aim to showcase the advancements and efficacy of our approach, particularly in bilingual image generation and fidelity to textual prompts.

\subsection{Machine Evaluation}

\paragraph{CLIP Model Evaluation.} Our CLIP model's performance is exemplary on both English and Chinese datasets, as evidenced by the zero-shot image-text retrieval results. The original CLIP model \citep{radford2021clip}, while establishing a foundational understanding, exhibits modest retrieval rates on the Flickr \citep{young2014flickr} and MSCOCO datasets \citep{lin2014coco}. This outcome highlights the inherent challenges associated with cross-lingual transfer learning. In contrast, AltCLIP \citep{chen2022altclip} and our enhanced CLIP model demonstrate significant improvements, with our model achieving the highest recall rates across most evaluation metrics. Particularly noteworthy is our model's performance in the Text $\rightarrow$ Image retrieval task on the Flickr-CN \citep{young2014flickr} and MSCOCO-CN datasets \citep{li2019coco-cn}, where it attains recall rates of 88.1\% and 69.7\% at R@1, respectively. These results indicate a robust alignment between textual prompts and visual content, underscoring the effectiveness of our tailored modifications in enhancing CLIP's cross-lingual performance. The results, presented in Table \ref{tabs:clip_combined}, demonstrate the potential of specialized models in handling diverse linguistic contexts within multimodal AI applications. The superior performance of our CLIP model, particularly in bilingual contexts, significantly bolsters the capabilities of the Taiyi-XL model. This enhancement allows for a more nuanced understanding of user-input prompts, leading to the generation of images that more accurately reflect the given prompts. The results affirm the importance of developing robust bilingual comprehension capabilities in models for advanced multimodal applications.

\begin{table*}[!h]
\centering
\resizebox{0.95\textwidth}{!}{%
\begin{tabular}{@{}lcccccccccccc@{}}
    \toprule 
    & \multicolumn{6}{c}{Flickr30K} & \multicolumn{6}{c}{MSCOCO} \\
    & \multicolumn{3}{c}{Image $\rightarrow$ Text} & \multicolumn{3}{c}{Text $\rightarrow$ Image} & \multicolumn{3}{c}{Image $\rightarrow$ Text} & \multicolumn{3}{c}{Text $\rightarrow$ Image} \\
    \cmidrule(lr){2-4} \cmidrule(lr){5-7} \cmidrule(lr){8-10} \cmidrule(lr){11-13}
    Model & R@1 & R@5 & R@10 & R@1 & R@5 & R@10 & R@1 & R@5 & R@10 & R@1 & R@5 & R@10 \\
    \midrule
    CLIP \citep{radford2021clip}  & 85.1 & 97.3 & 99.2 &  65.0 & 87.1 & 92.2 &  56.4 & 79.5 & 86.5 &  36.5 & 61.1 & 71.1 \\
    AltCLIP \citep{chen2022altclip} & 86.0 & 98.0 & 99.1 &  72.5 & 91.6 & 95.4 &  58.6 & 80.6 & 87.8 &  42.9 & 68.0 & 77.4 \\
    Our-CLIP                     & \textbf{88.4} & \textbf{98.8} & \textbf{99.9} &  \textbf{75.7} & \textbf{93.8} & \textbf{96.9} &  \textbf{61.2} & \textbf{84.8} & \textbf{90.3} &  \textbf{49.2} & \textbf{70.3} & \textbf{79.6}  \\
    \midrule
    & \multicolumn{6}{c}{Flickr30K-CN} & \multicolumn{6}{c}{MSCOCO-CN} \\
    & \multicolumn{3}{c}{Image $\rightarrow$ Text} & \multicolumn{3}{c}{Text $\rightarrow$ Image} & \multicolumn{3}{c}{Image $\rightarrow$ Text} & \multicolumn{3}{c}{Text $\rightarrow$ Image} \\
    \cmidrule(lr){2-4} \cmidrule(lr){5-7} \cmidrule(lr){8-10} \cmidrule(lr){11-13}
    CLIP~\citep{radford2021clip}  & 2.3  & 8.1  & 12.6 & 0    & 2.4  & 4.0  & 0.6  & 4.1  & 7.1  &  1.8  &  6.7 &  11.9 \\
    AltCLIP~\citep{chen2022altclip} & 69.8 & 89.9 & 94.7 & 84.8 & 97.4 & 98.8 & 63.9 & 87.2 & 93.9 & 62.8  & 88.8 &  95.5\\
    Our-CLIP                     & \textbf{73.2} & \textbf{90.3} & \textbf{96.5} & \textbf{88.1} & \textbf{98.2} & \textbf{99.1} & \textbf{66.0} & \textbf{91.1} & \textbf{96.6} & \textbf{69.7}  & \textbf{91.3} &  \textbf{96.8}  \\
    \bottomrule
\end{tabular}
}
\caption{Zero-shot image-text retrieval results on Flickr30K, MSCOCO, Flickr30K-CN, and MSCOCO-CN datasets. The best results are marked in \textbf{bold}.}
\label{tabs:clip_combined}
\end{table*}

\paragraph{Diffusion Model Evaluation.}Based on the data presented in Table \ref{table:models_comparison_bilingual}, a comprehensive analysis of the performance of various models in bilingual image generation tasks reveals significant insights. The evaluation metrics used for this analysis include CLIP Similarity (CLIP Sim), Inception Score (IS), and Fréchet Inception Distance (FID), which collectively offer a robust assessment of model performance in terms of image quality, diversity, and alignment with textual descriptions. In the English dataset (COCO), our Taiyi-XL model demonstrates superior performance across all metrics, notably achieving the highest CLIP Sim score, the highest IS, and the most favorable FID. These results indicate that Taiyi-XL not only generates images that are closely aligned with the given text prompts but also ensures high image quality and diversity. The model outperforms other contenders such as Alt-Diffusion, SD-v1.5, and SD-XL, highlighting its effectiveness in handling English language prompts in image generation tasks. Similarly, in the Chinese dataset (COCO-CN), Taiyi-XL again stands out, achieving the best results with a CLIP Sim score, IS and FID. Compared to other models like Taiyi-v0.1, Alt-Diffusion, and Pai-Diffusion, Taiyi-XL exhibits a remarkable ability to generate high-quality images that are well-aligned with Chinese textual descriptions. This performance underscores the model's robust bilingual capabilities, making it particularly suitable for applications requiring high-fidelity image generation from diverse linguistic inputs.

Overall, the results from both datasets affirm the efficacy of the Taiyi-XL model in bilingual image generation tasks. Its ability to consistently produce high-quality, diverse images that accurately reflect the content of both English and Chinese text prompts positions it as a leading solution in the field of multimodal AI applications. The superior performance of Taiyi-XL in these bilingual contexts highlights the potential of specialized models in navigating the complexities of different linguistic environments within image generation tasks.

\begin{table}[t]
\centering
\small
\begin{tabular}{lccc}
\toprule
Model & CLIP Sim($\uparrow$)  & FID($\downarrow$) & IS($\uparrow$) \\
\cmidrule(r){1-4}
\multicolumn{4}{c}{English Dataset (COCO)} \\
\midrule
Alt-Diffusion\citep{Ye2023AltDiffusionAM} & 0.220 & 27.600 & 31.577 \\
SD-v1.5\citep{Rombach_2022_CVPR_sd} & 0.225 & 25.342 & 32.876 \\
SD-XL\citep{podell2023sdxl} & 0.231 & 23.887 & 33.793 \\
Taiyi-XL & \textbf{0.254} & \textbf{22.543} & \textbf{35.465} \\
\midrule
\multicolumn{4}{c}{Chinese Dataset (COCO-CN)} \\
\cmidrule(r){1-4}
Taiyi-v0.1\citep{fengshenbang} & 0.197 & 69.226 & 21.060 \\
Alt-Diffusion\citep{Ye2023AltDiffusionAM} & 0.220 & 68.488 & 22.126 \\
Pai-Diffusion\citep{wang2023pai} & 0.196 & 72.572 & 19.145 \\
Taiyi-XL & \textbf{0.225} & \textbf{67.675} & \textbf{22.965} \\
\bottomrule
\end{tabular}
\caption{\label{table:models_comparison_bilingual}
Comparison of different models based on CLIP Sim, IS, and FID across English (COCO) and Chinese (COCO-CN) datasets. The best results are marked in \textbf{bold}.}
\end{table}


\subsection{Human Preference Evaluation}

\begin{figure*}[t]
    \centering
    \includegraphics[width=\linewidth]{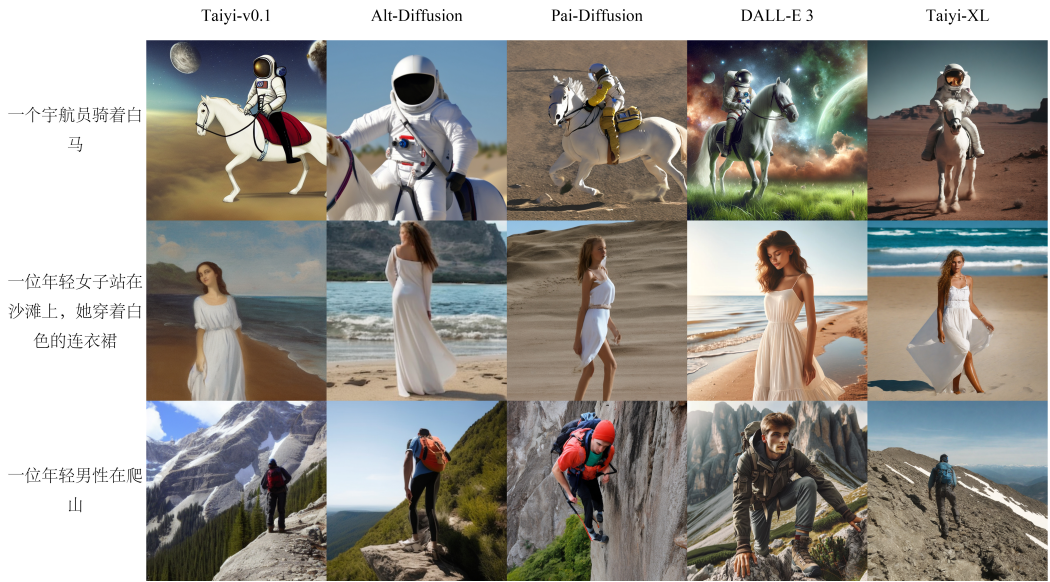}
    \caption{
    Comparison of Different Models in Chinese Text-to-Image Generation Performance.
    }
    \label{fig:zh_compare}
\end{figure*}

\begin{figure*}[t]
    \centering
    \includegraphics[width=\linewidth]{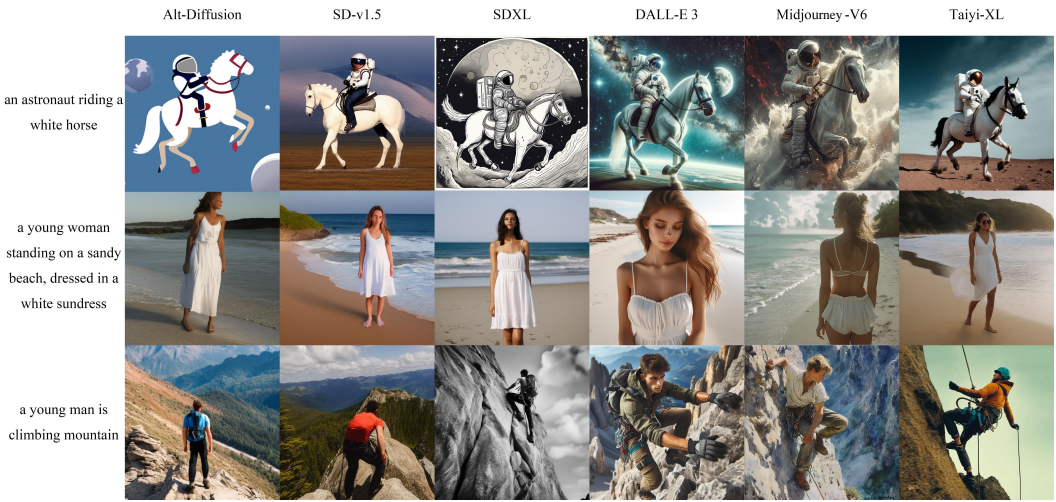}
    \caption{
    Comparison of Different Models in English Text-to-Image Generation Performance.
    }
    \label{fig:en_compare}
\end{figure*}

In our comprehensive analysis, as depicted in Figures \ref{fig:zh_compare} and \ref{fig:en_compare} showcasing the performance of various models in Chinese and English text-to-image generation, several key observations and conclusions have emerged. The XL versions of the models such as SD-XL and Taiyi-XL exhibit a significant improvement over the 1.5 versions such as SD-v1.5 and Alt-Diffusion, indicating advancements in the scale of model parameters, underlying algorithms and training methodologies. DALL-E 3, while occasionally producing overly vivid colors, stands out for its exceptional prompt-following capability, setting a high benchmark in generating images that closely align with the given textual descriptions. Our model, characterized by a photographic style, closely parallels the performance of Midjourney, particularly in its aesthetic appeal. However, a notable distinction lies in our model's enhanced support for bilingual (Chinese and English) text-to-image generation, a feature that is especially valuable in diverse linguistic contexts. This capability underscores the importance of language versatility in the realm of generative models.

The final conclusion drawn from this analysis is that while our model may not yet match the performance of commercial models, it significantly surpasses current bilingual open-source models. We attribute the gap with commercial models primarily to differences in the quantity, quality, and diversity of the image-text data used for training. Our model has been trained exclusively on copyright-compliant image-text data, highlighting the ongoing challenge of copyright issues in text-to-image and AI-generated content (AIGC) models. This aspect remains a critical factor in the development and refinement of generative models, underscoring the need for access to diverse and high-quality datasets while navigating the complexities of copyright constraints.

\begin{figure}[ht]
    \centering
    \begin{minipage}{.5\textwidth}
        \centering
        \includegraphics[width=\linewidth]{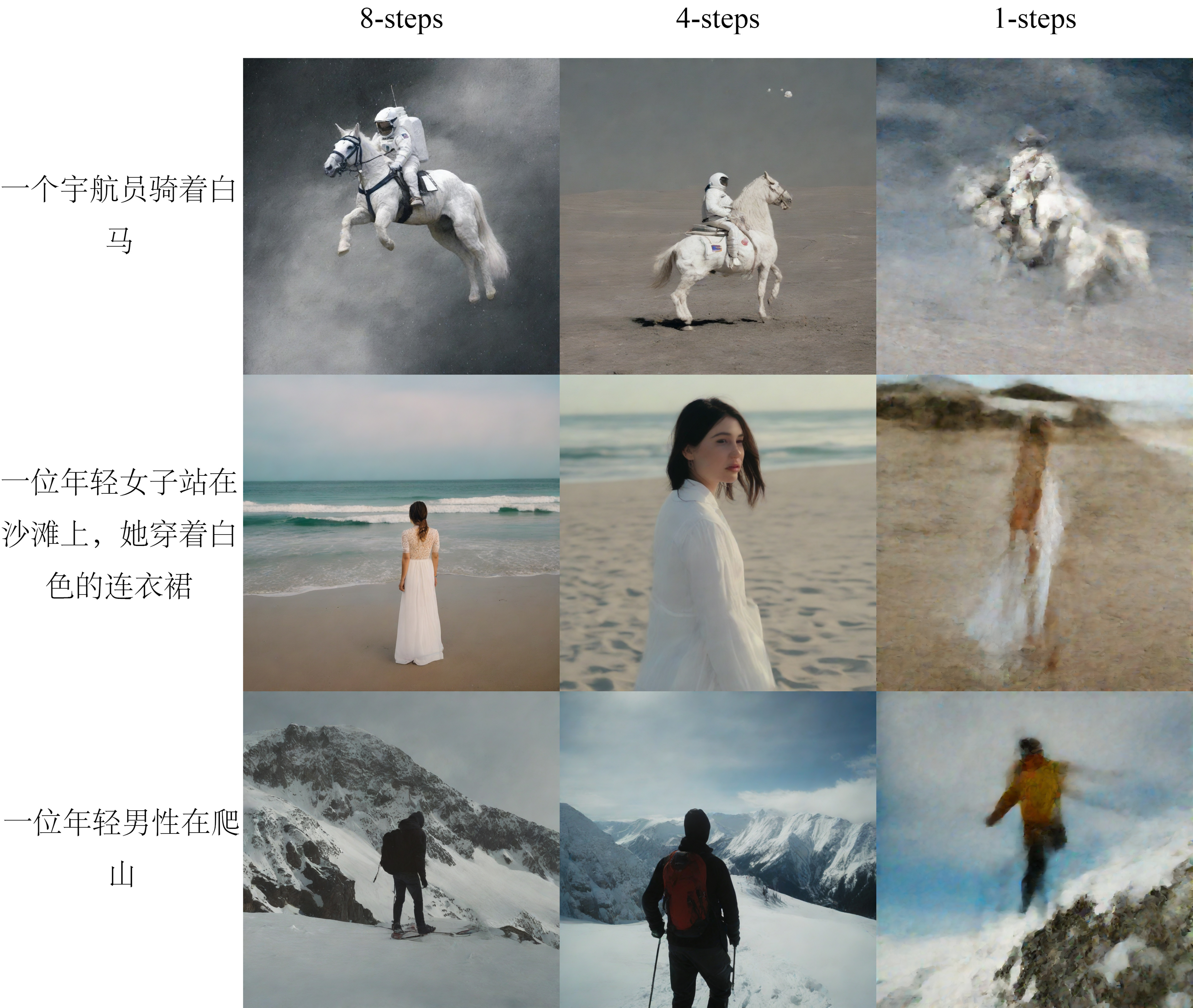}
    \end{minipage}%
    \begin{minipage}{.5\textwidth}
        \centering
        \includegraphics[width=\linewidth]{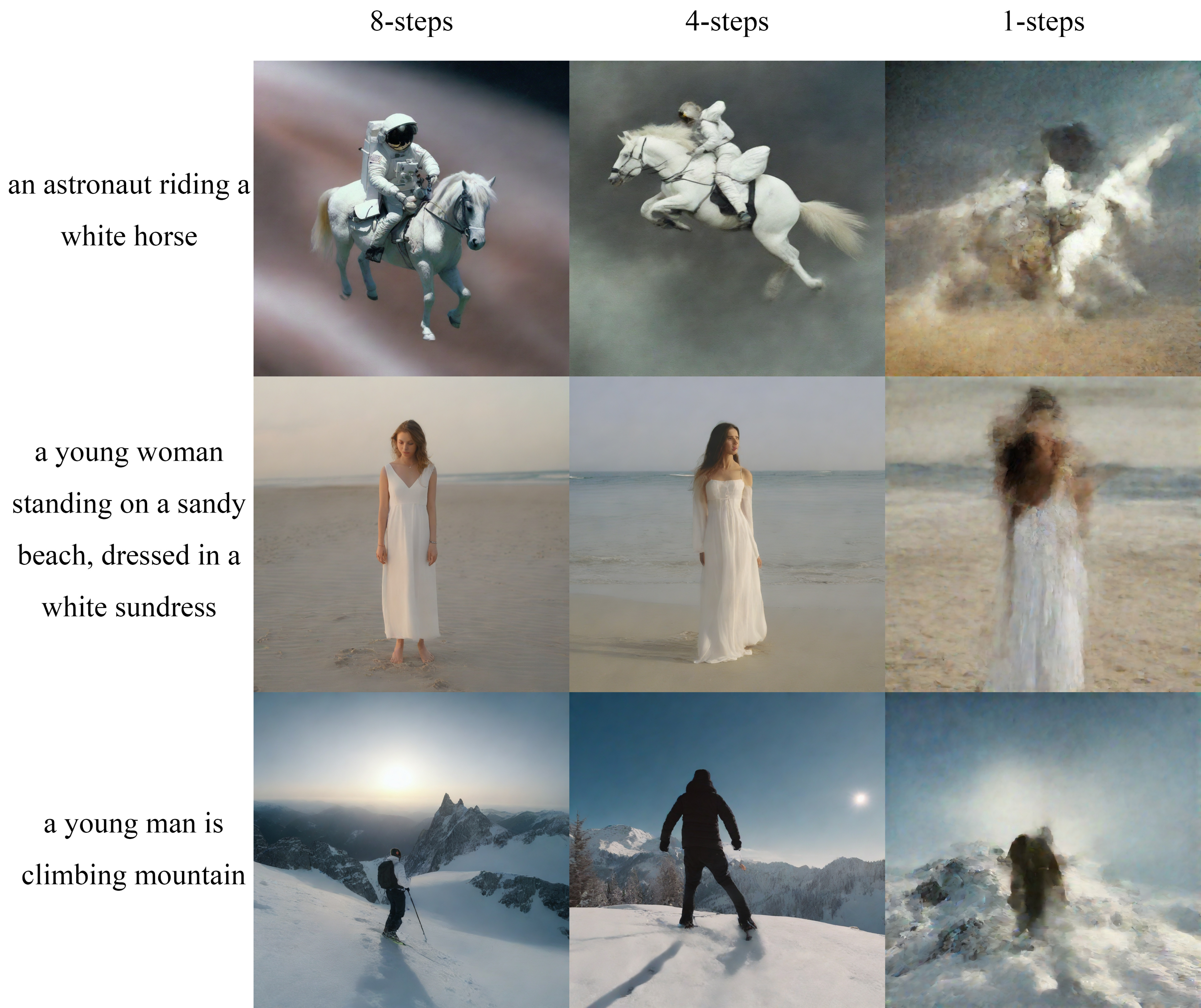}
    \end{minipage}
    \caption{Taiyi-XL generation examples with Latent Consistency Model}
    \label{fig:lcm}
\end{figure}

We also evaluated the impact of employing Latent Consistency Models (LCM) \citep{song2023cm,luo2023lcm,luo2023lcm_lora} to accelerate the image generation process. A notable observation \ref{fig:lcm} from these tests is the correlation between the reduction in inference steps and the consequent decline in image quality. Specifically, when the generation is constrained to a single step, the resulting images predominantly exhibit only basic outlines and lack finer details. However, extending the generation process to 8 steps ensures a considerably higher quality of the generated images. This finding suggests that while LCM can effectively speed up the generation process, a balance must be struck between the number of steps and the desired image quality. Maintaining a minimum number of steps, such as eight in our tests, appears to be crucial for preserving a satisfactory level of detail and overall image fidelity.

\section{Related Work}

\subsection{Advancements in Image Generation and Diffusion Models}
Recent years have seen substantial advancements in the field of text-to-image generation. This work diverges from traditional approaches such as Generative Adversarial Networks (GANs) \citep{goodfellow2014gan, arjovsky2017wgan}, Variational Autoencoders (VAEs) \citep{kingma2013vae}, Flow-based models \citep{rezende2015flow}, and autoregressive models \citep{ramesh2021dalle, ding2021cogview, ding2022cogview2}, focusing instead on the more advanced diffusion model. The evolution and refinement of diffusion theory and techniques \citep{vincent2011dsm, ho2020ddpm, song2020ddim, cao2022survey} have positioned the diffusion model as a leading technology in image generation. Noteworthy developments in this area include Dall-E 2 \citep{ramesh2022dalle2}, which utilizes a hierarchical approach for generating images based on textual descriptions with CLIP latents. Similarly, Imagen \citep{saharia2022imagen} and Deepfloyd-IF \citep{alex2023deep-floyd_IF} demonstrate the capability of diffusion models to produce photorealistic images from text, emphasizing deep language understanding. The latent diffusion model \citep{Rombach_2022_CVPR_sd}, encompassing works such as stable-diffusion-v1-5, stable-diffusion-2-1, and stable-diffusion-xl \citep{podell2023sdxl}, represents the forefront of this technology. These models primarily leverage the CLIP text model for textual feature extraction, integrating these features into the latent diffusion process to reduce computational overhead and memory requirements.

\subsection{Text-to-Image Models in Bilingual Context}
In response to the requirements of text-to-image generation in bilingual scenarios, especially in Chinese language, researchers have made significant contributions. initially, the CLIP text encoder is substituted with a Chinese-specific encoder, followed by pre-training for text-image matching on Chinese datasets. Key works in this domain include Taiyi-CLIP \citep{zhang2022fengshenbang}, Chinese-CLIP \citep{chineseclip}, and Alt-CLIP \citep{chen2022altclip}. Subsequently, the text encoder in stable diffusion is replaced, and further training on Chinese text-image datasets is conducted to enhance text-to-image generation capabilities. This leads to the development of Chinese versions of diffusion image generation models, such as Taiyi-diffusion \citep{zhang2022fengshenbang}, Alt-diffusion \citep{Ye2023AltDiffusionAM} and Pai-diffusion\citep{wang2023pai}. However, it is noteworthy that replacing the CLIP text encoder can result in the loss of English language capabilities in the model, and the training process can be resource-intensive.

\subsection{The Role of Text-Image Datasets}
Datasets are pivotal in both text-image matching and text-to-image generation. Traditional image caption datasets like COCO \citep{lin2014coco} and Flickr \citep{young2014flickr} in English, and COCO-CN \citep{li2019coco-cn} and Flickr-CN \citep{li2016fickr-cn} in Chinese, provide a foundational training base but are limited in size, generally below one million entries. Consequently, web-crawled datasets such as Laion\citep{schuhmann2021laion} (primarily in English) and Wukong\citep{gu2022wukong} (primarily in Chinese) have emerged as more critical data sources for training diffusion text-to-image models, boasting sizes of up to 100 million or even 5 billion.

\section{Conclusion}

Our research demonstrates the profound impact of integrating bilingual support into text-to-image models, significantly advancing multimodal research in Chinese contexts. The development of Taiyi-CLIP and Taiyi-XL models, with their expanded vocabulary and position encoding, marks a notable advancement in image-text retrieval and image generation. These models lay the foundation for future innovations in bilingual multimodal studies. Additionally, the use of large vision-language models to enrich text prompts has led to more accurate and detailed image generation, aligning closely with user intent. This approach underscores the importance of accurate and complex language understanding in text-to-image generation. As we continue to make our findings and models open-sourced, we invite collaboration and further exploration, contributing to a more inclusive and linguistically diverse future in artificial intelligence research.

\bibliography{iclr2021_conference}
\bibliographystyle{iclr2021_conference}

\newpage
\appendix
\vspace{2em}

\end{document}